\pdfoutput=1
\documentclass[11pt]{article}

\usepackage{EMNLP2023}
\usepackage{adjustbox}
\usepackage{amsmath}
\usepackage{amssymb}

\usepackage{times}
\usepackage{latexsym}

\newcommand{\crossm}{$\times$}
\newcommand{\tick}{$\checkmark$}
\newcommand{\quotes}[1]{``#1''}
\usepackage{float}
\usepackage{tabularx}

\usepackage[T1]{fontenc}
\usepackage[utf8]{inputenc}

\usepackage{microtype}
\usepackage{enumerate}
\usepackage{inconsolata}

\usepackage{graphicx}
\usepackage{multirow}
\usepackage{makecell}
\usepackage{tablefootnote}

\usepackage[T1]{fontenc}
\usepackage[utf8]{inputenc}

\usepackage{microtype}

\usepackage{inconsolata}

\newcommand{\titledigi}{HeySQuAD}
\newcommand{\digi}{HeySQuAD}
\newcommand{\digih}{HeySQuAD$_h$}
\newcommand{\digim}{HeySQuAD$_m$}
\newcommand{\squad}{SQuAD}

\title{\titledigi: A Spoken Question Answering Dataset}

\author{
  Yijing Wu \And Sai Krishna Rallabandi \And Ravisutha Srinivasamurthy \AND
 Parag Pravin Dakle \And Alolika Gon \And Preethi Raghavan \AND
  \normalfont{Fidelity Investments} \\
  \texttt{\{yijing.wu, saiKrishna.rallabandi, ravisutha.srinivasamurthy,}\\
  \texttt{paragpravin.dakle, alolika.gon, preethi.raghavan\}@fmr.com} \\
  }

\begin{document}
\maketitle
\begin{abstract}
%Human-spoken questions are critical to evaluating the performance of spoken question answering (SQA) systems that serve several real-world use cases including digital assistants. We present a new large-scale community-shared SQA dataset, \digi \, that consists of 76k human-spoken questions and 97k machine-generated questions and corresponding textual answers derived from the SQuAD QA dataset. 

%The goal of \digi \ is to measure the ability of machines to understand noisy spoken questions and answer the questions accurately. To this end, we run extensive benchmarks on the human-spoken and machine-generated questions to quantify the differences in noise from both sources and its subsequent impact on the model and answering accuracy. Importantly, for the task of SQA, where we want to answer human-spoken questions, we observe that training using the transcribed human-spoken and original SQuAD questions leads to significant improvements (12.51\%) over training using only the original \squad\ textual questions. 
Spoken question answering (SQA) systems are critical for digital assistants and other real-world use cases, but evaluating their performance is a challenge due to the importance of human-spoken questions. This study presents a new large-scale community-shared SQA dataset called \digi , which includes 76k human-spoken questions, 97k machine-generated questions, and their corresponding textual answers from the SQuAD QA dataset. Our goal is to measure the ability of machines to accurately understand noisy spoken questions and provide reliable answers. Through extensive testing, we demonstrate that training with transcribed human-spoken and original SQuAD questions leads to a significant improvement (12.51\%) in answering human-spoken questions compared to training with only the original SQuAD textual questions. Moreover, evaluating with a higher-quality transcription can lead to a further improvement of 2.03\%. This research has significant implications for the development of SQA systems and their ability to meet the needs of users in real-world scenarios.
\end{abstract}
\section{Introduction}\label{intro}
%Spoken language question answering (SQA)\cite{spoken_squad, contextualized_attention_knowledge_transfer_for_sqa,MRDnet_for_sqa,knowledge_distillation_for_sqa,self-supervised_dialoglearning_for_sqa}  has become increasingly pervasive in our day-to-day lives as we frequently talk to digital assistants on devices including our phones, and smart watches, asking them to retrieve factual information or take actions to help us. For e.g., \quotes{Hey \textit{Agent}, when does the World Cup start?}
%This may include answering several factual  questions, checking on the weather, setting reminders, playing songs, etc. 
Spoken language question answering (SQA)\cite{spoken_squad, contextualized_attention_knowledge_transfer_for_sqa,MRDnet_for_sqa,knowledge_distillation_for_sqa,self-supervised_dialoglearning_for_sqa} is a rapidly growing field with diverse applications in our daily lives, including digital assistants on mobile devices, smartwatches, and other platforms. These systems enable us to perform various tasks or retrieve factual information by speaking to them. However, the complexity and variability of human speech, coupled with noisy transcriptions and the contextual dependence of natural language, make the problem of SQA challenging.

\begin{table}
\centering
\begin{tabular}{|p{7.1cm}|}
\hline
\textbf{Context}: This Main Building, and the library collection, was entirely destroyed ... The hall housed multiple classrooms and science labs needed for early research at the university.  \\

\textbf{Original question}: In what year was the Main Building at Notre Dame razed in a fire?\\

\textbf{Transcribed human-spoken question}: a might yet was the main building in another name raised in a fire \\

 \textbf{Transcribed machine-generated question}: in what year was the main building at notre dame raised in a fire \\

\textbf{Answer}:  \{\quotes{text}: \quotes{1879}, \quotes{answer\_start}: 90\} \\

\hline
\end{tabular}
\caption{An example instance from \digi\ consisting of context, original \squad\ question, ASR transciptions of human-spoken and machine-generated questions, and the answer.}\label{table:data_example}
\end{table}

%insert figure with examples here
%insert citations in the intro.
%highlight that getting human voice to train is costly 
At the heart of this challenge lies the need for an accurate semantic understanding of both speech and text data, which is followed by the provision of precise answers to the posed questions. This involves automatic speech recognition (ASR) and textual question answering (QA), both of which require addressing the aforementioned challenges (demonstrated in  Table \ref{table:data_example}).

To develop models that can handle transcription noise in SQA, we introduce a novel dataset called \digi, derived from the widely used \squad \cite{squad} reading comprehension datasets. We use an ASR model\footnote{\url{https://huggingface.co/facebook/s2t-small-librispeech-asr}} to obtain transcriptions of both human-spoken and machine-generated questions and create two datasets, \digih  \ and \digim \ for the SQA experiments. \\
\textbf{\squad\ train and dev}: in the \squad\ 1.1 format. \\
\textbf{\digih \ train and dev}: Contains \squad\ context and transcribed human-spoken questions in the \squad\ 1.1 format. \\
\textbf{\digim \ train and dev}: Contains \squad\ context and transcribed machine-generated questions in \squad\ 1.1 format.

All three datasets consist of 48849 training and 1002 dev examples. An example instance of \digih, \digim, and \squad\ is shown in Table \ref{table:data_example}. 
%\digi\ consists of 76k human-spoken questions and 97k machine-generated questions and corresponding textual answers derived from SQuAD. 
%The main goal of \digi\ is to help develop models that can effectively handle transcription noise and answer the spoken question accurately. 
Our main contributions are as follows.

(1) We release the first large-scale community-shared dataset\footnote{\url{https://github.com/yijingjoanna/HeySQuAD}}, consisting of 76k human-spoken audio questions and textual answers, as well as 97k machine-generated audio questions and their ASR transcripts. This dataset offers a valuable resource for various tasks, including ASR or SQA model improvement, audio-text multi-modal models, and better matching/retrieval efficiency. Our dataset enables extensive benchmarking and model development, propelling community efforts towards improved spoken language question answering models (Section \ref{section:dataset_description}). More directly, \digi \ can be used to enable SQA on the SQuAD/SQuAD-like datasets  and used to pretrain SQA models for other domains/datasets.

(2) We introduce an SQA leaderboard based on the \digi \ dataset, providing extensive benchmarks using state-of-the-art models for question answering across \digi\ transcribed questions and original \squad \ questions. The experiments leading up to these benchmarks allow us to make key observations that impact the training mechanisms and performance of SQA models (Table \ref{table:QA_results_machine_human}, and Table \ref{table:QA_results_whisper}). 
\begin{enumerate}[i]
    \item We observe that training using \squad\ textual questions and transcribed human-spoken questions by LibriSpeech achieves an f1 score of 90.18\%, which is a significant improvement (12.51\%) over training using only the \squad\ textual questions (77.67\%) when evaluated on human-spoken questions by LibriSpeech. 
    \item Even training using machine-generated questions (low-cost) is significantly better (by 11.34\%) than training with original \squad, which sets the stage for easily obtainable machine-generated questions to be used for training more accurate models for SQA.
    \item In addition, we include \digih \ transcribed by Whipser, which achieves better transcription quality and run various experiments to show the effect of different transcription qualities on the SQA. 
\end{enumerate}
The value of our contributions lies in providing key insights derived from a comprehensive dataset and leaderboard for the SQA research community, enabling the development of more accurate and reliable SQA models.
%(3) Define a SQA challenge around \digi\ to encourage community-efforts towards working on this important, practically-relevant task.

%perhaps move this to related work.
%Spoken question answering (SQA) is a type of NLP task that involves generating a written or spoken response to a spoken question. SQA systems can either be open domain or closed domain - with the questions being limited to a specified context in spoken or written form. Since SQA systems are designed to understand and respond to questions that are posed in a natural and conversational manner, they can be used in a variety of applications, including virtual assistants, customer service chatbots, and educational software. Therefore, building such systems has attracted a lot of interest in the recent years. 

%To address these challenges, SQA systems use a combination of techniques and technologies, such as acoustic modeling, language modeling, syntactic analysis, semantic analysis, and structured databases. Despite significant progress in SQA, there are still many open research questions and challenges that need to be addressed, such as improving the accuracy and robustness of speech recognition and natural language understanding, handling out-of-vocabulary words and rare events, and developing more effective evaluation metrics and benchmarks for SQA systems.

\section{Related Work}\label{section:related_works}
\textbf{Spoken Question Answering}
% \cite{odsqa_adversarialadaptation} present an approach using adversarial learning to alleviate this gap. 

Spoken Question Answering (SQA) \cite{spoken_squad,odsqa,odsqa_adversarialadaptation, odsqa} involves a set of approaches to answer a spoken question appropriately. There are two types of SQA systems: in context SQA systems \cite{spoken_squad, contextualized_attention_knowledge_transfer_for_sqa,MRDnet_for_sqa,knowledge_distillation_for_sqa,self-supervised_dialoglearning_for_sqa} and open domain SQA systems. While open domain SQA systems are built with the aim to answer questions without context, in context SQA systems typically take the form of Machine Reading Comprehension (MRC) \cite{mrc_multimodalattention} where a passage accompanies the question.  Models employing the pipeline architecture have highlighted the effects of errors arising from the ASR component \cite{spoken_squad}. \citealp{sqa_contextualizedwordrep} explored the use of sub-word strategy while \citealp{speechbert} utilized the BERT model to learn joint audio-text features with the aim of mitigating the errors from ASR. However, such techniques do not explicitly consider the mapping between ASR transcript and the corresponding human transcription. 
In our work, we present experiments with
models explicitly aimed at exploiting this mapping
as well, by attempting to directly learn the said mapping in a translation fashion. In \citealp{selfsupervised_contrastive_replearning}, the authors present a method to learn representations
using contrastive learning and show improved performance on three SQA datasets. While the augmentation strategies presented may implicitly contribute to improved representations learned, it is unclear how they explicitly improve the performance of an SQA system. In this paper, we
employ contrastive learning as a regularize, thereby explicitly regulating the final objective. 

\textbf{SQA Datasets} \citealp{spoken_squad} release a large scale dataset for SQA with 37k and 5.4k question-answer pairs in training and test sets respectively. The dataset was created by converting the passage part of \squad\ dataset into speech by using Text-to-Speech. While this dataset paves the way for research into SQA, the spoken part of the dataset contains just the passages. The noise characteristics captured in the dataset reflect those of synthetic data. Our work presents a dataset of synthetic and human-recorded questions, characterizing a larger and close-to real-life variability. \citealp{spoken_coqa} release a conversational question-answering dataset that consists of question-answering pairs from 7 domains. While this dataset is closer to the real-life scenario (since it captures a conversation with dialog turns), it still uses synthetic data (and we were unable to independently verify this dataset's existence even after our best efforts). Our dataset reflects the real-world scenario more closely since it captures human variability while asking the questions. \citealp{odsqa} present a small dataset with reading comprehension recorded from TOEFL. While this dataset contains real human speech data, it is created based on just 6 test candidates and hence is very limited. Our dataset contains around 80K human-recorded questions.  \citealp{sdqa} present the SD-QA dataset derived from the TyDi-QA dataset with human-spoken questions. In English, 1k questions in the dev set and 1k in the test set are recorded in 11 English dialects by native speakers, and in total, SD-QA contains more than 68k questions in 5 languages and 24 dialects. Though this dataset contains questions spoken in different languages and dialects, only questions in the dev and test set are included, and the number of questions is very limited. To support the development of SLU models that take advantage of pre-trained speech representations, SLUE\cite{shon-etal-2023-slue} releases models for many tasks. This includes, for each task, (i) curated annotations for a relatively small fine-tuning set, (ii) a reproducible pipeline (speech recognizer + text model) along with end-to-end baseline models and evaluation metrics, and (iii) baseline model performance in various types of systems to facilitate easy comparisons. Additionally, they examine the impact of speech recognition accuracy on pipeline models' performance by utilizing over 20 publicly available speech recognition models.
\begin{table*}
    \centering
    \resizebox{\linewidth}{!}{
        \begin{tabular}{|*{9}{c|}} 
            \hline
            % Header (next 5 lines)
            \textbf{Dataset Name} & 
            \multicolumn{2}{c|}{\textbf{Human Spoken Context} }& 
            \multicolumn{2}{c|}{\textbf{Human Spoken Questions}} & 
            \multicolumn{2}{c|}{\textbf{Machine Spoken Context}} & 
            \multicolumn{2}{c|}{\textbf{Machine Spoken Questions}}\\
            \cline{2-9}
            
            % Subheader
            & Yes? & Size & Yes? & Size & Yes? & Size & Yes? & Size \\

            \hline
            % Content
            % & & & & & & & & \\
            % Dataset Name                            | Human Spoken Context                          | Human Spoken Questions                      | Machine Spoken Context               | Machine Spoken Question
            \large{\squad 1.1} \cite{squad}                  & \crossm   & \thead{T: 0 \\[0.2em] D: 0}        & \crossm & \thead{T: 0 \\[0.2em] D: 0}              & \crossm & \thead{T: 0 \\[0.2em] D: 0}       & \crossm & \thead{T: 0 \\[0.2em] D: 0} \\[0.5em]
            \large{Spoken \squad} \cite{spoken_squad}        & \crossm   & \thead{T: 0 \\[0.2em] D: 0}               & \crossm & \thead{T: 0 \\[0.2em] D: 0}              & \tick & \thead{T: 37k \\[0.2em] D: 5.4k}    & \crossm & \thead{T: 0 \\[0.2em] D: 0} \\[0.5em]
            \large{SCQA} \cite{spoken_coqa}                  & \tick     & \thead{T: 0 \\[0.2em] D: 500}             & \tick & \thead{T: 0 \\[0.2em] D: 500}              & \tick & \thead{T: 7.1k \\[0.2em] D: 0}      & \tick   & \thead{T: 7.1k \\[0.2em] D: 0} \\[0.5em]
            \large{TOEFL} \cite{toefl}                       & \tick     & \thead{T: 963 \\[0.2em] D: 124 \\[0.2em] TT: 122}& \crossm & \thead{T: 0 \\[0.2em] D: 0}              & \crossm & \thead{T: 0 \\[0.2em] D: 0}       & \crossm & \thead{T: 0 \\[0.2em] D: 0} \\[0.5em]
            \large{SD-QA} \cite{sdqa}                        & \crossm   & \thead{T: 0 \\[0.2em] D : 0}              & \tick & \thead{T: 0 \\[0.2em] D: 1k*11}            & \crossm & \thead{T: 0 \\[0.2em] D: 0}       & \crossm & \thead{T: 0 \\[0.2em] D: 0} \\[0.5em]
            \large{SLUE-SQA-5}\cite{shon-etal-2023-slue} & \textbf{\tick}   & \thead{\textbf{T: 15k} \\[0.2em] \textbf{D: 1.6k} \\[0.2em] \textbf{TT: 2k}}            & \textbf{\tick} & \thead{T: 46k \\[0.2em] D: 1.9k \\[0.2em] TT: 2.4k}   & \crossm & \thead{T: 0 \\[0.2em] D: 0}       & \crossm & \thead{T: 0 \\[0.2em] D: 0} \\
            \large{\textbf{\digi\ (ours)}}                    & \crossm   & \thead{T: 0 \\[0.2em] D: 0}               & \textbf{\tick} & \thead{\textbf{T: 72k} \\[0.2em] \textbf{D: 4k}}   & \crossm & \thead{T: 0 \\[0.2em] D: 0}       & \textbf{\tick} & \thead{\textbf{T: 87k} \\[0.2em] \textbf{D: 10k}} \\
         \hline
        \end{tabular}
    }
    \caption{Different spoken QA datasets are compared along with their \textbf{t}rain (referred to as \textbf{T}), \textbf{d}ev (\textbf{D}) and \textbf{t}est (\textbf{TT}) sizes. \digi \ is the first large-scale community-shared dataset containing 76k human-spoken and 97k machine-generated questions, which is unique across these datasets. For the SCQA dataset, the train and test size are the numbers of conversations. For SD-QA, \quotes{1k*11} represents 1k questions in 11 English dialects.}
    \label{table:dataset_comparison}
\end{table*}

% Most SQA systems \cite{contextualized_attention_knowledge_transfer_for_sqa,MRDnet_for_sqa,knowledge_distillation_for_sqa,self-supervised_dialoglearning_for_sqa} are based on an integrated pipeline architecture i.e they combine Automatic Speech Recognition(ASR) system (which converts the spoken content into text) with Text based Question Answering (TQA) system that generates answers to textual questions by conditioning on the passage.

\section{\digi \ Dataset  Description}\label{section:dataset_description}
%\subsection{Corpus Description}
% TODO: add annotators info

% The answer to each question is always a span in the context. SQuAD1.1 contains 107,785 question-answer pairs on 536 Wikipedia articles and is partitioned into a training set (80\%), a development set (10\%), and a test set (10\%). SQuAD2.0 combines the 100k questions in SQuAD1.1 with over 50k unanswerable questions written adversarially by crowd workers to look similar to answerable ones. (need to cite squad dataset paper here).
We created HeySQuAD by recording human-spoken and machine-generated questions in the SQuAD dataset along with the ASR transcriptions.
\subsection{Annotation details}
%i change passage to context
%The annotators were recruited after approval from an internal ethics board. 
We employed a total of 12 speakers - 5 females and 7 males - who are all native-English speakers with no accents. For synthetic audio questions, we use the Amazon Polly service and choose single speaker \quotes{Joanna} with \quotes{neural} engine to output mp3 files with a sample rate of 16KHz. 
All questions corresponding to a \squad\ context were shown to the annotator one after another. The annotator was required to record one question at a time; thus, sentences of the same
context were spoken by the same speaker. We believe this
is closer to a real-life scenario where the user has a multi-turn interaction with the system. 

%Q - How many speakers does SpokenQA use? 12 speakers (7 male, 5female? diversity)
%Q - Number of tokens comparison? 
%            Total number of tokens  Average #tokens in a question
%Original    492k                    10.09
%Machine     518k                    10.61
%Human       540k                    11.07
%Q - Length of utterances (do humans take more time to speak?)
%Q - More pauses?

%\textbf{Machine-generated synthetic audio questions. } 

%The original, human-spoken and machine-generated training dataset contains 48849 original, 48849 human-spoken and 48849 machine-generated questions, respectively. The original, human-spoken and machine-generated dev dataset contains 1002 original, 1002 human-spoken and 1002 machine-generated questions, respectively.
\subsection{Transcriptions of Spoken Questions}
Experiments on SQA require train and dev datasets with \squad\ 
 textual questions, transcribed human-spoken questions, and transcribed machine-generated questions.  We use an automatic speech recognition ASR model (facebook/s2t-small-librispeech-asr) \cite{wang2020fairseqs2t} to obtain the transcription of human-spoken and machine-generated questions and create the human-spoken \digih \ and machine-generated \digim \ datasets (Section \ref{intro}) for the SQA experiments.

\subsection{Dataset Characteristics}\label{subsec:dataset_chara}
%Why are we doing this WER analysis? Write down the motivation first.

In order to quantify the noise in the transcribed human-spoken and machine-generated questions, we analyze the Word Error Rate (WER). Lower WER indicates better transcription quality. The WER for transcribed human-spoken and machine-generated questions using ASR model (facebook/s2t-small-librispeech-asr) in Table \ref{table:WER_error} shows that there are more errors in transcibed human-spoken \digih \ than machine-generated  \digim.

\begin{table}
\centering
\resizebox{0.7\columnwidth}{!}{%
\begin{tabular}{|c|c|c|}

\cline{1-3} \textbf{Dataset} &  \textbf{WER} of train & \textbf{WER} of dev \\ 
\cline{1-3}
 \digih   & 0.3423  & 0.2722 \\ 
 \digim  & 0.2087   & 0.1961 \\
\cline{1-3}
\end{tabular}%
}
\caption{WER of transcribed human-spoken and machine-generated questions by LibriSpeech ASR.}\label{table:WER_error}
\end{table}

Typically, errors in the transcribed human-spoken utterances is known to be a result of errors from the ASR model and also because of noises such sampling errors from microphone, background noise, speaker speed, pronunciation and etc. in the recorded audio \cite{WER_error}. In comparison, the errors in the transcribed machine-generated questions result from the ASR model while transcribing audio to text, and TTS models while generating synthetic audio.

We now specifically investigate the errors in \digi\ by sampling 100 random examples from the dev set where we have the \squad\ textual questions, transcribed human-spoken and transcribed machine-generated questions. Examples and quantities of different types of errors are shown in Table \ref{table:WER_error_examples} in Appendix. 

From this result, we can infer that the human pronunciation and recording quality vs TTS machine-generated synthetic audio quality is the main reason impacting the transcription quality and results in the lower WER of transcribed machine-generated questions than human-spoken questions.

We analyze the WER of \digih\ and \digim\ dev set for each question type in Table \ref{table:WER_by_type}, Appendix \ref{appendix:wer}. We also provide the part-of-speech (PoS) tags in Table \ref{table:PoS_analysis}, Appendix \ref{appendix:POS}.

 %human pronounciation vs machine-generated TTS are main reasons that impact the transcription quality/WER

%Following is one example of the transcription error:
%\begin{itemize}\label{}
%   \item original question: Which country \textbf{refused} to content to changes in the Treaty of \textbf{Lisbon 2007}?
%    \item transcribed human-spoken question: which country \textbf{refuse} to content to changes in the treaty of \textbf{lips on two thousand and seven}
%    \item transcribed machine-generated question: which country refused to content to changes in the treaty of lisbon \textbf{two thousand seven}
%\end{itemize}

% Different Datasets Comparison
\renewcommand\theadfont{\large}

%\section{Challenge Formulation}
%\input{sections/task_description}
%\input{sections/evaluation}

\section{Models}
\textbf{Spoken Query Answering with ASR (QuASR):}
Let $Q_i$ be the input question and $S_i$ be ${Q_i}$ in spoken form. Also let $P_i$ be the passage in textual form. We pass $S_i$ through an ASR system to obtain the transcription $T_i$. Also consider $A_i$ as the answer to the question. QuASR learns the conditional distribution P($A_I$ / $T_i$). Our model can be summarized by the following set of equations:

\begin{equation} \label{eq1}
\begin{aligned}
T_i ={}& \textbf{H}^{ASR}(S_i) \\
A_i ={}& \textbf{QuASR}(T_i) \\
\end{aligned}
\end{equation}
\textbf{Loss Functions:} As the output for our models is the span of the answers, we use cross entropy loss function to penalize the errors made in prediction. 
\textbf{Evaluation Metrics:}
The SQA models return a text answer based on the transcribed questions and contexts. Since the task involves retrieval of a span of text, we compute the exact match (EM) and F1 scores between the predicted text answer and the ground truth text answer to gauge the performance of models. See Appendix \ref{appendix:em_f1}. 

\section{Experiments}
We run experiments on six QA models: bert-base-uncased, bert-large-uncased, roberta-base, roberta-large, albert-base-v2 and albert-large-v2 \cite{transformers} and fine-tune in four methods:\\
\textbf{\squad}: fine-tuned on the \squad\ train. \\
\textbf{\digih\ / \digim}: fine-tuned on the \digih\ train or \digim\ train. \\
\textbf{\squad\ and \digih\ / \digim}: fine-tuned on both \squad\ and \digih\ train, or fine-tuned on both \squad\ and \digim\ train.\\
\textbf{\squad\ then \digih\ / \digim}:fine-tuned first on \squad, and then fine-tuned on \digih\ or \digim\ train.   \\
Each model was trained for 2 epochs with learning rate of 3e-5 and maximum length of tokens in the sequence being 512.

%For our JoiQA system, we have employed a model that learns joint distribution between $S_i$ and $P_i$. The encoders of both $S_i$ and $T_i$ are realized using bidirectional LSTMs. We have used 256 as the hidden dimensions for both these encoders. To adjust for the difference in temporal resolution of $S_i$ and $T_i$, we downsample the frames of $S_i$. 
%\subsection{Hyperparameters}

\section{Results}\label{section:results}
\begin{table*}[t]
\centering
\resizebox{15cm}{!}{%
\begin{tabular}{{|c|c|l|cc|cc|cc|}}
\hline
\textbf{QA Model}&& \textbf{Train Dataset}& \multicolumn{2}{c|}{\textbf{\small{\squad  }}}& \multicolumn{2}{c|}{\textbf{\small{\digim } }}& \multicolumn{2}{c|}{\textbf{\small{\digih }  }}\\ 
&& & \multicolumn{2}{c|}{\textbf{ dev}}& \multicolumn{2}{c|}{\textbf{dev}}& \multicolumn{2}{c|}{\textbf{dev}}\\ 
\hline
&&& F1&EM&F1&EM&F1&EM\\
\hline
\multirow{7}{*}{\makecell{bert-large\\-uncased} } &a& \squad\    train& 90.01&	81.74&     77.38	& 64.57& 75.29	& 62.67        \\
&b  & \digih \ train& 88.95	& 81.14& 86.70&	78.84& 84.88	&77.25 \\
&c  & \squad\  and   \digih\ train& 88.85 &	80.94& 85.64	& 77.74 & 83.82	& 75.65  \\
&d   & \squad\  then \digih \ train & 88.72	& 80.94& 85.07	& 77.64& 83.88 &	75.95\\
&e & \digim\ train& 89.81& 82.03& 86.51& 78.64& 84.56& 78.45\\
&f  & \squad\ and   \digim\ train& 89.25& 81.43& 86.32& 77.64& 83.23& 73.75\\
&g& \squad\ then \digim\ train & 89.60& 81.74& 85.83& 77.44& 84.23& 75.25\\
\hline
\multirow{7}{*}{\makecell{roberta-large}} &a& \squad\    train                                & 93.71 &	86.52 & 80.78	&67.26	&\textbf{77.67}	& 63.87             \\
                                 &b    & \digih \ train                        & 92.71	&84.83 &	92.05&	85.22	&89.14	&81.93 \\
                                &c   & \squad\  and   \digih \ train         & 93.60	&86.72&	92.11&	85.62	&\textbf{90.18}	&83.73                     \\
                                 &d    & \squad\   then \digih \ train & 93.11	&86.52	&91.12	&84.03	&88.52	&81.23               \\
                                 &e    & \digim\ train                        & 93.41                        & 86.32                        &  92.46 &  85.72 & \textbf{89.01}	& 82.03 \\
                                  &f   & \squad\ and   \digim\ train         & 93.41                         & 86.52                        & 92.11                        & 85.92                       & 89.13                        & 81.93                        \\
                               &g      & \squad\  then \digim\ train & 93.23                        & 86.32                        & 91.13                        & 83.73                       & 89.01                        & 82.03                        \\
                                  \hline
\multirow{7}{*}{\makecell{albert-large-v2}}  &a   & \squad\   train                                 & 91.49&	84.03	&65.05&	41.91&	62.01	&38.82\\
&b    & \digih \ train                        & 90.67	&82.33	&88.03	&80.03	&87.18	&79.04     \\
                                 &c    & \squad\  and   \digih \ train         & 91.50	&84.23	&89.02	&81.43	&87.77	&79.74               \\
                                 &d    & \squad\   then \digih \ train & 91.51	&84.73	&88.73	&80.93	&87.33	&79.64                    \\
                                 &e    & \digim\ train                        & 90.98                        & 82.73                        & 89.09                        & 81.53                        & 86.78	&78.84\\
                                &f     & \squad\ and   \digim\ train         & 91.60                         & 84.43                        & 89.00                          & 81.13                       & 85.85                        & 77.24                        \\
                                &g    & \squad\  then \digim\ train & 92.51                        & 85.12                        & 89.29                        & 80.63                       & 86.94                        & 78.34                        \\
\hline
\end{tabular}%
}
\caption{QA accuracy results (\%) using bert-large-uncased, roberta-large, albert-large-v2 QA models fine-tuned on \digih, \digim \ and \squad\  train dataset; numbers in bold show when evaluated on the human-spoken \digih\ dev set, the best F1: 90.18\% is obtained by fine-tuning on the \squad\ and human-spoken \digih, which is 12.51\% higher than fine-tuned only on \squad\  (77.67\%), and fine-tuning on even machine-generated \digim\ train which is low-cost achieves 89.01\%.}\label{table:QA_results_machine_human}
\end{table*}

\begin{table*}[t]
\centering
\resizebox{15cm}{!}{%
\begin{tabular}{{|c|c|c|c|c|c|}}
\hline
\textbf{QA Model}&& \textbf{Train Dataset}& \textbf{Val Dataset} & \textbf{F1}&\textbf{EM}\\ 
\hline
\multirow{10}{*}{\makecell{roberta-large}} &a& \squad\    train      and   \digih \ train by LibriSpeech &          \digih \ dev by LibriSpeech                 & \textbf{90.18} &	83.73 	\\ 
                                 &b    &\squad\    train      and   \digih \ train by LibriSpeech &          \digih \ dev by whisper                 & 92.19&	85.33 	\\ 
                                &c   & \squad\    train      and   \digih \ train by whisper &          \digih \ dev by LibriSpeech                 &  84.75&76.35	\\ 
                                 &d  &  \squad\    train      and   \digih \ train by whisper &          \digih \ dev by whisper              &  91.21 & 83.23 	\\ 
                                 &e  &  \squad\    train    &          \digih \ dev by LibriSpeech                 & 77.67	& 63.87	\\ 
                                 &f  &  \squad\    train   &          \digih \ dev by whisper                 & 92.15	& 84.63 	\\ 
                                 &g  &  \digih \ train by LibriSpeech    &          \digih \ dev by LibriSpeech                 &  89.14& 81.93	  	\\ 
                                 &h  &  \digih \ train by LibriSpeech   &          \digih \ dev by whisper                 & \textbf{92.21} 	&  84.43 	\\ 
                                 &i  &  \digih \ train by whisper    &          \digih \ dev by LibriSpeech                 & 84.70 	&  76.85	\\ 
                                 &j  &  \digih \ train by whisper   &          \digih \ dev by whisper                 & 90.49 	&  82.83 	\\ 
                                  \hline
\end{tabular}%
}
\caption{QA accuracy results (\%) using roberta-large QA models fine-tuned on \digih and \squad\ train dataset using two different ASR models: LibriSpeech and Whisper.}\label{table:QA_results_whisper}
\end{table*}

%In this section, we present results in two categories - (1) Results of our models on Spoken Query Answering and (2) Results of our models on Textual Question Answering corresponding to the same dataset used for Spoken QA. These results can be seen in table \ref{table_results}. 

We present results on SQA task using QA models fine-tuned on SQuAD, machine-generated \digim\ and human-spoken \digih, evaluated on the \squad, \digim \ and \digih \ dev set in Table \ref{table:QA_results_machine_human}. When evaluated on the human-spoken \digih\ dev which is the closest to the real life use case of SQA, the best F1 score 90.18\% is obtained by using roberta-large QA model fine-tuned on the \squad\ and \digih, which is 12.51\% higher than roberta-large QA model fine-tuned only on the \squad\ (77.67\%). In addition, using roberta-large QA model fine-tuned on even \digim\ which is low-cost achieves 89.01\%. We make the following observations:
%enumerate
%for human dev
%add titles that tell us what is the point of this
\textbf{Models fine-tuned on the human-spoken \digim\ or machine-generated \digih\ perform significantly better than fine-tuned only on the \squad}, comparing Row a, b and e; Column \digih \ dev. 
    
\textbf{Models fine-tuned on the human-spoken \digih \ train have higher scores than models fine-tuned on the machine-generated \digim \ train, but not significantly,} differences in F1 scores are less than 0.4\% by comparing Row b and e ; Column \digih \ dev, although the WER in Table \ref{table:WER_error} implies less noisy \digim\ than \digih \ train.

\textbf{Adding \squad \ to training has no significant improvement} when evaluating on the \digih dev (Row b to d for fine-tuning with human-spoken \digih \ and Row e to g for machine-generated \digim; Column \digih \ dev. Differences in F1 scores are less than 1.66\%, which implies training using only \digim \ or \digih \ train is sufficient with no need of original textual questions.\\
We notice that for each model, accuracy scores when evaluated on \squad\ dev is higher than evaluated on machine-generated \digim\ dev, higher than evaluated on human-spoken \digih dev. This implies getting higher-quality transcription of the dev set improves SQA. \\
We also include SQA results using bert-base-uncased, roberta-base, albert-base-v2 QA models in Appendix Table \ref{table:QA_results_base_machine_human}. The performance of model for different question types is tabulated in table \ref{table:micro_wer} in the Appendix. Error analysis on the SQA are kept in Appendix \ref{appendix:error_analysis}.

%Comparison of different model BERT,RoBERTa,ALBERT QA model fine-tuned on SQuAD1.1 with 2 epochs, learning rate = 3e-5 in Table \ref{table_baseline1}. 
%\section{Error Analysis}
%\input{sections/error_analysis}

% \input{sections/discussions.tex}
\section{Improvements by better ASR model}

% Say that the micro analysis is in agreement with macro analysis
\iffalse

\begin{table}
\centering
\resizebox{0.7\columnwidth}{!}{%
\begin{tabular}{|c|c|c|}

\cline{1-3} ASR Transcription Error & WER & BLEU  \\ 
\cline{1-3}
 \digih \ Train by LibriSpeech  & 0.3422 & 0.5303  \\ 
 \digih \ Dev LibriSpeech & 0.2722  &  0.5914\\ 
  \digih \ Train by whisper  & 0.1088 & 0.8317  \\ 
 \digih \ Dev by whisper & 0.0471  & 0.8475 \\ 
\cline{1-3}
\end{tabular}%
}
\caption{WER and BLEU of transcribed human-spoken questions by LibriSpeech and Whisper ASR models.}\label{table:WER_BLEU_whisper}
\end{table}
\fi
\begin{table}
\centering
\resizebox{\columnwidth}{!}{%
\begin{tabular}{|c|c|c|c|}

\cline{1-4} \textbf{Dataset} & \textbf{ASR model} & \textbf{WER} of train & \textbf{WER} of dev  \\ 
\cline{1-4}
\multirow{2}{*}{\digih}&  LibriSpeech  & 0.3422 & 0.2722   \\ 
& Whisper  & 0.1088 & 0.0471  \\ 
\cline{1-4}
\end{tabular}%
}
\caption{WER of transcribed human-spoken questions by LibriSpeech and Whisper.}\label{table:WER_BLEU_whisper}
\end{table}

% Continue with different types of transcription errors  (who/what analysis) [waiting for some data]

% Discussions
 %The primary goal of \digi\ is to develop a dataset for SQA that allows us to build models that help understand noisy spoken questions and answer them accurately. 

%should we say discussions in bold as a title on 497? Or here. 

% is that good? (line 147 of this doc)

In this section, we use the OpenAI whisper-large-v2 ASR model \cite{radford2022whisper} (Whisper) to obtain a higher quality transcription of the human-spoken questions in \digih \ compared to ASR model facebook/s2tsmall-librispeech-asr (LibriSpeech) which is used in all previous sections. Table \ref{table:WER_BLEU_whisper} shows WER using these two ASR models.

We intend to investigate the impact of different ASR models on the SQA task. We use roberta-large model fine-tuned on \squad \ and \digih \ human-spoken question by LibriSpeech and Whisper, evaluated on \digih \ by LibriSpeech and Whisper respectively. Results of SQA are shown in table \ref{table:QA_results_whisper}. F1 score of 90.18 is the best SQA score using LibriSpeech ASR on the human-spoken questions in \digih. When using Whisper ASR on the human-spoken questions in \digih, the best F1 score is improved to 92.21\% (row h) which shows an increment of 2.03\%, achieved by finetuning with \digih \ train by LibriSpeech.

We find the better transcription quality in the dev dataset results in higher accuracy of SQA when the QA model and training dataset are the same. Especially a difference of 14.48\% in row e and f when fine-tuning with \squad \ train. Adding \squad \ train to \digih \ train only makes a difference less than 1.04\% in the F1 score, comparing row a and g, b and h, c and i, and d and j respectively; which means, in real use case of training SQA model, it is not necessary to have the gold transcription of questions in the train dataset.\\
When evaluated on \digih \ by Whisper, comparing scores between b, d, f, h and j, the difference of F1 scores are less than 1.72 \%. On the contrary, when evaluated on \digih \ by LibriSpeech, adding LibriSpeech transcribed \digih \ train dataset to the \squad \ train improves 12.51 \% by comparing row a and e, and using \digih \ by LibriSpeech improves SQA F1 score by 5.43 \% and 4.44\% compared with using \digih \ by Whisper, see row a and c, row g and i respectively.\\
\textbf{Observations:} When training an SQA model using transcribed questions, the transcription quality of the dev set plays a crucial role in SQA accuracy. If the dev set has high transcription quality, such as \digih\ dev by Whisper, the choice of training with or without the transcribed \digih dataset, or using different transcriptions of \digih (by Whisper or LibriSpeech), does not significantly impact the SQA accuracy (with differences less than 1.72\%).\\
However, when the transcription quality is less satisfactory, such as the \digih\ dev by LibriSpeech, a substantial difference of 12.51\% in SQA accuracy F1 score is observed when fine-tuning the RoBERTa-large QA model with varying training datasets. These findings underscore the importance of transcription quality in the development of SQA systems, which has practical implications for various industries that rely on natural language processing technologies.\label{sec:whisper}

%\section{Improvement by LLM}
%\input{sections/LLM}

%\section{Limitations and Potential Risks}
%\input{sections/limitations.tex}

\section{Conclusion}
% We introduce a new large-scale SQA dataset, \digi, comprising 76k human-spoken and 97k machine-generated audio and transcribed questions, along with their corresponding textual answers derived from SQuAD. Our extensive benchmarks on both human-spoken and machine-generated questions, using two different ASR models provide insights into the differences in noise from these sources and their subsequent impact on model performance and answer accuracy. The findings emphasize the practical implications of our research for the development of SQA systems in various industries that rely on natural language processing technologies.

We introduce a new large-scale dataset, \digi, for Spoken Question Answering (SQA). This dataset consists of 76,000 human-spoken and 97,000 machine-generated audio questions, along with their corresponding textual answers derived from SQuAD. Our extensive benchmarks on both human-spoken and machine-generated questions, using two different ASR models, provide insights into the differences in noise from these sources and their subsequent impact on model performance and answer accuracy. The findings emphasize the practical implications of our research for the development of spoken question answering systems in various industries that rely on natural language processing technologies.

\section*{Ethics Statement}
We present HeySQuAD, which includes both human-spoken and machine-generated audio and transcribed questions. This will help future researchers study various effects of variability in human speech and design approaches to alleviate spurious transcriptions. 

\bibliography{main}  
\bibliographystyle{acl_natbib}

\appendix
\section{ASR error analysis}\label{appendix:asr_error}
We show examples and quantities of the errors of proper nouns/technical words, and errors from the same pronunciations in the 100 random examples of transcribed Human-spoken and machine-generated questions. The rest of the transcription errors, marked as \quotes{Others}, can arise from any source such as ASR model error or sampling error while recording or from bad microphone quality or because of background noise etc, and it is difficult to identify the sources of an error when it occurs. In addition, out of 100 examples, 44 contain proper nouns and technical words, which means in human-spoken questions, 70.45\% of these proper nouns and technical words are wrong, and in machine-generated questions, 65.91\% are wrong. 
\begin{table*}
\centering
\begin{tabular}{|c|c|c|c|}
\hline
\textbf{Error Category }  & \makecell{\textbf{Example Ground Truth -}\\ \textbf{Transcription} } & \makecell{\textbf{Machine-generated} \\ \textbf{questions}} & \makecell{\textbf{Human-spoken} \\ \textbf{questions}} \\ \hline
\makecell{Proper nouns\\Technical words}  & Wahhabi - wad be & 29\% & 31\% \\ 
\hline
\makecell{Error from the\\ same pronunciation} & \makecell{1907 - ninteen and seven,\\ policymakers - policy makers,\\ site - sight} & 17\% & 16\% \\ 
\hline
Others      & In what year - in white year & 25\% & 44\%\\ 
\hline
\end{tabular}
\caption{Examples and quantities of errors in the transcription of 100 random human-spoken and machine-generated questions, showing the higher WER for human-spoken than machine-generated questions mainly results from the \quotes{others} category which is a combination of ASR model error, sampling error while recording human voice or generating synthetic audio, and etc.}\label{table:WER_error_examples}
\end{table*}

\section{Evaluation Metrics}\label{appendix:em_f1}
we employ two types of evaluation metrics to gauge the performance of models: computing the exact match (EM) and F1 scores between the predicted text answer and the ground truth text answer. For instance, if the predicted text answer is exactly the same as the ground-truth text answer the EM score is 1 and F1 is 1. However, when there is a mismatch between them, EM score is 0 where F1 scores are between 0 and 1, based on the precision and recall. Precision is the percentage of words in the predicted answer existing in the ground-truth answer, while the recall is the percentage of words in the ground-truth answer also appearing in the predicted answer.

\section{WER analysis}\label{appendix:wer}

WER analysis of \digim\ dev and \digih\ dev grouped by question types is shown in the table \ref{table:WER_by_type}.  The question type \quotes{others} contains questions in which we are unable to find any question word. For each question type, \digim\ dev have better transcribed questions than \digih. Especially, we notice that there are more \digih\ dev questions (10.88\%) in \quotes{others} compared to \digim\ dev (8.89\%) and \squad\  dev (4.19\%). We also note that the WER of \digih\ in \quotes{others} is significantly higher than the \digim. The performance of model for different question types is tabulated in table \ref{table:micro_wer}. The highest WER is seen for the question type \quotes{how} (ignoring the WER for \quotes{others}) and it can be seen that the model performance degrades significantly when the model is not trained on transcribed data. Hence, training model on transcribed data (either generated by machine or spoken by humans) make the model robust.

\begin{table*}[th]
\centering
\begin{tabular}{|c|c|cc|cc|}
\hline
\multicolumn{1}{|c|}{\textbf{Question}} & \multicolumn{1}{c|}{\textbf{\squad\ dev}} &\multicolumn{2}{c|}{\textbf{\digim\ dev}} & \multicolumn{2}{c|}{\textbf{\digih\ dev}} \\\cline{2-6}
\textbf{ Type}    &  \textbf{Percentage}    &  \textbf{Percentage}     & \textbf{WER }       & \textbf{Percentage}    & \textbf{WER}       \\
 \hline
who              &8.38\%    & 8.18\%                  & 0.2320     & 7.98\%                 & 0.2555    \\
what          &63.97\%       & 62.28\%                 & 0.1830     & 56.69\%                & 0.2394    \\
where         &    2.79\%   & 2.69\%                  & 0.2350     & 3.19\%                 & 0.2414    \\
when        & 7.29\%        & 6.79\%                  & 0.1607     & 7.09\%                 & 0.2301    \\
why         &1.90\%         & 1.80\%                  & 0.1257     & 2.79\%                 & 0.2740    \\
how        &   11.48\%       & 11.38\%                 & 0.2010     & 11.38\%                & 0.2533    \\
others     &  4.19\%        & 6.89\%                  & 0.3016     & 10.88\%                & 0.5111   \\
overall &100\% &100\% &0.1961 &100\% &0.2722\\
\hline
\end{tabular}
\caption{WER analysis of \digim\ dev and \digih\ dev grouped by question types}\label{table:WER_by_type}
\end{table*}

\begin{table*}
\centering
\resizebox{15cm}{!}{%
\begin{tabular}{|c|c|c|cc|cc|cc|}
\hline
\makecell{\textbf{Question}  \\ \textbf{Type}} & \makecell{\textbf{Question} \\ \textbf{Percentage}} & \textbf{WER} & \multicolumn{2}{c|}{\makecell{\textbf{Fine-tuned on}  \\ \textbf{\squad\  train}}} & \multicolumn{2}{c|}{\makecell{\textbf{Fine-tuned on}   \\\textbf{\squad\ } \\\textbf{and \digih\  train}} }& \multicolumn{2}{c|}{\makecell{\textbf{Fine-tuned on} \\ \textbf{ \digim \ train}} }\\
\hline
                             &    &                                      & F1                            & EM                           & F1                         & EM                        & F1                         & EM                        \\
\hline
who                  & 7.98\%     &  0.2555        & 81.89& 72.5& 96.69& 93.75                     &           94.48	&92.5\\
what                 & 56.69\%   & 0.2394         & 80.44& 68.31& 90.45& 83.3                      &             90.42 &	83.63\\
where                & 3.19\%    & 0.2414          & 70.32& 43.75& 92.86& 87.5                      &                88.48 &	78.13                                \\
when                 & 7.09\%    & 0.2301          & 82.53& 83.24& 98.7& 97.18                     &                   95.65	& 92.96                       \\
why                  & 2.79\%     & 0.2740         & 75.92& 39.28& 80.65& 64.29                     &                 77.44 & 	57.14                       \\
how                  & 11.38\%     &0.2533        & 76.46& 52.63& 89.65& 73.4                      &                 90.85	& 83.33                      \\
others               & 10.88\%     &0.5111        & 60.88& 52.29& 80.71& 73.39                     &                  74.53	&65.14                         \\
overall (macro-avg)             & 100\%&  0.2722 & 77.67& 64.58& 90.19& 82.52                     &                  89.01	&82.04                    \\
\hline
\end{tabular}%
}
\caption{roberta-large QA model macro-average accuracy scores on \digih\ dev of different question types}
\label{table:micro_wer}
\end{table*}

\section{POS analysis}\label{appendix:POS}

PoS tagging of 1002 \squad\ questions, transcribed machine-generated questions and human-spoken questions using spaCy \quotes{en\_core\_web\_sm} model \cite{spacy} shown .

\begin{table}
\centering
\resizebox{6.7cm}{!}{%
\begin{tabular}{|c|c|c|c|}
\hline
\makecell{\textbf{PoS} \\\textbf{Tagging}}  & \textbf{Original} & \makecell{\textbf{\digim \ }} & \makecell{\textbf{\digih \ }} \\ \hline
DET   & 297      & 329     & 319   \\ \hline
NOUN  & 3124     & 3281    & 3344  \\ \hline
AUX   & 857      & 819     & 817   \\ \hline
PROPN & 8        & 11      & 15    \\ \hline
VERB  & 1336     & 1367    & 1391  \\ \hline
ADP   & 1429     & 1449    & 1531  \\ \hline
ADV   & 443      & 435     & 469   \\ \hline
ADJ   & 1042     & 1062    & 1094  \\ \hline
PRON  & 684      & 720     & 771   \\ \hline
NUM   & 145      & 314     & 313   \\ \hline
INTJ  & 2        & 3       & 4     \\ \hline
PART  & 171      & 170     & 180   \\ \hline
CCONJ & 114      & 159     & 174   \\ \hline
SCONJ & 78       & 75      & 85    \\ \hline
X     & 5        & 9       & 19    \\ \hline
SYM   & 0        & 0       & 2     \\ \hline
\end{tabular}%
}
\caption{PoS tagging of 1002 \squad\ questions, transcribed machine-generated questions and human-spoken questions.}\label{table:PoS_analysis}
\end{table}

\section{SQA Results}
The following table shows SQA performance on the bert-base-uncased, roberta-base, alberr-base-v2 QA models.
\begin{table*}[t]
%\begin{tabular}{|c|c|cc|cc|p{0.6cm}p{0.6cm}|}
\centering
\resizebox{15cm}{!}{%
\begin{tabular}{{|c|c|p{6.7cm}|cc|cc|cc|}}
\hline
\textbf{QA Model  }                    &        & \textbf{Train Dataset  }                                        & \multicolumn{2}{c|}{\textbf{\small{\squad\  }}}            & \multicolumn{2}{c|}{\textbf{\small{\digim\ } }}            & \multicolumn{2}{c|}{\textbf{\small{\digih\ }  }}             \\ 
                                &    &                                       & \multicolumn{2}{c|}{\textbf{ dev}}            & \multicolumn{2}{c|}{\textbf{dev}}            & \multicolumn{2}{c|}{\textbf{dev}}             \\ 

\hline
                          &           &                                                        & F1&EM&F1&EM&F1&EM                                       \\
                                     \hline
\multirow{7}{*}{\makecell{bert\\-base\\-uncased} } & a &\squad\  train & 87.31	&78.64	&71.82&	55.49&	67.34	&50.80                       \\
                                    &b & \digih\ train                        & 85.25&	75.95	&82.36	&73.55&	80.89	&71.16                     \\
                                  &c   & \squad\  and   \digih\ train         & 86.33&	77.25&	82.39	&72.65&	81.13&	71.66                       \\
                               &d      & \squad\  then \digih\ train & 85.62	&75.85	&82.12&	71.66	&80.71&	70.06                 \\
                                   &e  & \digim\ train                        & 86.34&	77.15&	83.44	&74.05	&80.18&	70.96 \\
                                   &f  & \squad\  and   \digim\ train         &86.04&	77.15&	82.33&	72.85&	80.02	&69.76\\
                                  &g  & \squad\  then \digim\ train & 86.88	&77.84	&82.53&	72.45	&80.65&	70.66\\
\hline
\multirow{7}{*}{\makecell{roberta\\-base}} &a & \squad\    train                                & 90.14	&82.53	&83.94&	74.65	&\textbf{79.94}&	69.86 \\
                                 &b    & \digih\ train                        & 91.13	&83.73	&88.84	&81.93&	\textbf{87.34}	&79.84\\
                                  &c   & \squad\  and   \digih\ train         & 91.07&	84.03	&88.16	&80.73	&86.87	&79.14                     \\
                               &d      & \squad\   then \digih\ train & 89.89	&82.33	&86.95&	79.14	&85.93	&78.14                       \\
                                 &e    & \digim\ train                        &90.83	&83.93	&88.75	&81.8	&\textbf{86.00}&	78.74 \\
                                &f   & \squad\  and   \digim\ train         &90.87	&84.23&	88.28&	80.83	&85.13	&76.84                    \\
                                 &g  & \squad\   then \digim\ train & 89.25	&82.03	&87.45	&79.04	&84.8&	76.74              \\
                                 \hline
\multirow{7}{*}{\makecell{albert\\-base-v2}}  &a   & \squad\    train                                & 88.93&	81.33	&70.10	&50.89&	66.52	&48.00\\
                                 &b    & \digih\ train                        & 89.19	&81.63	&86.69	&79.14	&85.28&	77.34\\
                                &c     & \squad\  and   \digih\ train         & 90.10	&82.83	&86.43	&78.94&	84.92	&77.44\\
                                &d     & \squad\   then \digih\ train & 89.24&	81.73&	85.40	&77.54	&83.89&	76.04                     \\
                                 &e    & \digim\ train                        & 89.97	&82.63&	87.26&	79.64&	85.28&	77.34  \\
                                 &f    & \squad \  and   \digim\ train         & 88.77&	80.73&	83.98&	75.44&	82.58	&72.35              \\
                                 &g   & \squad  \   then \digim\ &88.93	&81.33	&70.06	&50.89&	66.52&	48.00                 \\
\hline
\end{tabular}%
}
\caption{QA accuracy results using bert-base-uncased, roberta-base, alberr-base-v2 QA models fine-tuned on \digih, \digim\ and \squad\  train dataset. The bold numebers show that when evaluated on the \digih\ dev set, the best accuracy score (F1: 87.34\%) is obtained by using roberta-base QA model fine-tuned on the human-spoken \digih\ dataset, which is 7.40\% higher than fine-tuned only on the \squad\ dataset (79.94\%). In addition, using roberta-large QA model fine-tuned on even machine-generated \digim\ train which is low-cost achieves 86.00\%. }\label{table:QA_results_base_machine_human}
\end{table*}

\section{Error Analysis }\label{appendix:error_analysis}
\newcommand{\q}{$q$}
\newcommand{\qhat}{$\hat{q}$}

% Error analysis starts here
% ===============================================================
In order to quantify the challenges involved in dealing with variability and complexity of human speech, the best model (roberta-large) from the section \ref{section:results} was tested on transcribed human-spoken \digih \ dev set. 

One of the important aspect the \digih dev set is the amount of transcription error. If we consider clean textual questions from \squad\ set (denoted as \q) and human speech transcribed question from \digih\ set (denoted as \qhat), we get two sets of possibilities: (a) When $\hat{q} \approx q$ and (b) When $\hat{q} \neq q$. To measure the similarity between \q\ and \qhat\, we use a sentence transformer \cite{sbert} to get the embeddings and then compute the cosine similarity between \q\ and \qhat. When the similarity between the \q\ and \qhat\ is greater than 0.9, we consider the transcription to be free of transcription errors ($\hat{q} \approx q$). In this setting, the model trained on the clean dataset (referred to as \squad\ model) is expected to perform on par with the model trained on transcribed human-spoken dataset (referred to as \digih\ model). Table \ref{table:micro_same_questions} shows the performance difference ($\sim$7\%) when the questions are very similar. As the questions are very similar, the error form these models are mostly \textit{model errors}.

When the similarity between \q\ and \qhat\ is less than 0.9, we assume that the transcribed text contains significant transcription errors ($\hat{q} \neq q$). In this setting, the only difference between model trained on \squad\ and the model trained on \digih\ is that the \digih\ model is aware of \textit{transcription errors}. From Table \ref{table:micro_same_questions}, it can be inferred that the performance of the \squad\ model degrades significantly ($\approx$16\% deficit compared to \digih\ model's F1) in presence of transcription errors.

\iftrue
\begin{table}
    \centering
        \begin{tabular}{|c|c|c|} 
            \hline
            \textbf{Model} & \textbf{$q \approx \hat{q}$} & \textbf{$q \neq \hat{q}$} \\
             & \textbf{F1} & \textbf{F1}\\
            \hline
            \digih\ Model & 94.36 & 87.66 \\
            \squad Model & 87.23 & 71.90\\
            \hline
            Difference in F1 & 7.12 & \textbf{15.76} \\
         \hline
        \end{tabular}
    \caption{When transcribed questions are very similar to clean textual questions ($q \approx \hat{q}$), we expect the model trained on clean set to perform as well as model trained on transcribed set. And when different ($q \neq \hat{q}$) (mostly due to transcription error), we observe that the performance of \digih\ model is ($\sim$16\%) more than \squad\ model's performance}
    \label{table:micro_same_questions}
\end{table}
\fi
\textbf{Discussion. }
From the above analysis, it is clear that the \textit{transcription errors} have huge impact on the model performance. The types of transcription errors have been  discussed in section \ref{section:dataset_description}. Additional analysis of how WER for different question types (who/when/why etc.) affects model performance is presented in the the Appendix \ref{appendix:wer}.

\end{document}